\def\year{2020}\relax
\documentclass[letterpaper]{article} % DO NOT CHANGE THIS
\usepackage{aaai20}  % DO NOT CHANGE THIS
\usepackage{times}  % DO NOT CHANGE THIS
\usepackage{helvet} % DO NOT CHANGE THIS
\usepackage{courier}  % DO NOT CHANGE THIS
\usepackage[hyphens]{url}  % DO NOT CHANGE THIS
\usepackage{graphicx} % DO NOT CHANGE THIS
\usepackage{graphicx}  % DO NOT CHANGE THIS
\usepackage{times}
\usepackage{latexsym}

\usepackage{bm}
\usepackage{graphicx}

\usepackage{makecell}

\usepackage{booktabs}
\usepackage{amsfonts}
\usepackage{soul}
\usepackage{enumitem}

\usepackage{soulpos}
\usepackage{url}
\usepackage{adjustbox}
\usepackage{subcaption}
\usepackage{ragged2e}

\usepackage{comment}

\usepackage{multirow}
\usepackage{booktabs}
\usepackage{array}
\newcommand{\citet}[1]{\citeauthor{#1}~\shortcite{#1}}
\usepackage{tikz}
\usepackage{pgfplots}
\pgfplotsset{width=5cm,compat=1.3, legend style={at={(0.40,0.40)},anchor=north west}}
\usepackage{amsmath}

\usepackage{algorithm,algorithmic}

\DeclareMathOperator*{\argmax}{argmax}
\DeclareMathOperator*{\argmin}{argmin}
\usepackage{xcolor}
\usepackage{colortbl}
\definecolor{myorange}{RGB}{193, 128, 0}
\definecolor{myblue}{RGB}{5, 96, 175}
\definecolor{mygreen}{RGB}{34,139,34}
\definecolor{mygrey}{RGB}{118,118,118}

\definecolor{grey}{rgb}{0.8,0.8,0.8}

\usepackage{url}
\title{Is BERT Really Robust? A Strong Baseline for Natural Language Attack \\on Text Classification and Entailment}
\author{Di Jin,\textsuperscript{\rm 1}$^{*}$ Zhijing Jin,\textsuperscript{\rm 2}\thanks{Equal Contribution. Order determined by swapping that in the previous paper at https://arxiv.org/abs/1901.11333. % \cite{jin2019imat}
} Joey Tianyi Zhou,\textsuperscript{\rm 3} Peter Szolovits\textsuperscript{\rm 1}\\  % All authors must be in the same font size and format. Use \Large and \textbf to achieve this result when breaking a line
\textsuperscript{\rm 1}Computer Science \& Artificial Intelligence Laboratory, MIT\\
\textsuperscript{\rm 2}University of Hong Kong\\
\textsuperscript{\rm 3}A*STAR, Singapore\\
jindi15@mit.edu,
zhijing.jin@connect.hku.hk,
zhouty@ihpc.a-star.edu.sg,
psz@mit.edu % email address must be in roman text type, not monospace or sans serif
}

\begin{document}

\maketitle

\begin{abstract}
 Machine learning algorithms are often vulnerable to adversarial examples that have imperceptible alterations from the original counterparts but can fool the state-of-the-art models. It is helpful to evaluate or even improve the robustness of these models by exposing the maliciously crafted adversarial examples. In this paper, we present \textsc{\textbf{TextFooler}}, a simple but strong baseline to generate adversarial text. By applying it to two fundamental natural language tasks, text classification and textual entailment, we successfully attacked three target models, including the powerful pre-trained BERT, and the widely used convolutional and recurrent neural networks. We demonstrate three advantages of this framework: (1) effective---it outperforms previous attacks by success rate and perturbation rate, (2) utility-preserving---it preserves semantic content, grammaticality, and correct types classified by humans, and (3) efficient---it generates adversarial text with computational complexity linear to the text length.\footnote{The code, pre-trained target models, and test examples are available at \url{https://github.com/jind11/TextFooler}.}
\end{abstract}

\section{Introduction}

In the last decade, Machine Learning (ML) models have achieved remarkable success in various tasks such as classification, regression and decision making. However, recently they have been found vulnerable to adversarial examples that are legitimate inputs altered by small and often imperceptible perturbations \cite{kurakin2016adversarial,DBLP:journals/corr/KurakinGB16a,papernot2017practical,zhao2017generating}. These carefully curated examples are correctly classified by a human observer but can fool a target model, raising serious concerns regarding the security and integrity of existing ML algorithms. On the other hand, it is showed that robustness and generalization of ML models can be improved by crafting high-quality adversaries and including them in the training data \cite{43405}.

\begin{figure}[!t]
    \centering
    \includegraphics[width= \columnwidth]{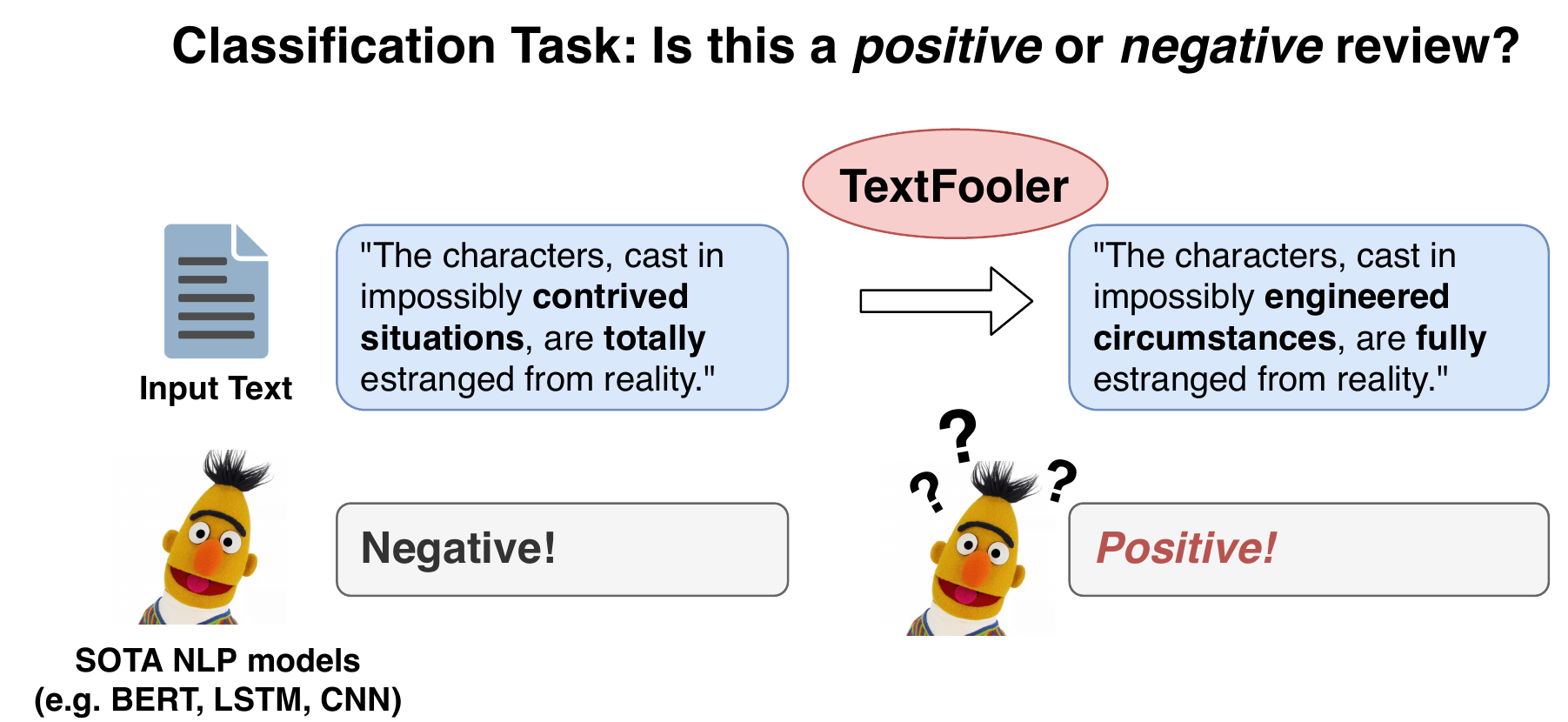}
    \caption{Our model TextFooler slightly change the input text but completely altered the prediction result.}
    \label{fig:intro-ex}
\end{figure}
While existing works on adversarial examples have obtained success in the image and speech domains \cite{szegedy2013intriguing,carlini2018audio}, it is still challenging to deal with text data due to its discrete nature. Formally, besides the ability to fool the target models, outputs of a natural language attacking system should also meet three key utility-preserving properties: (1) human prediction consistency---prediction by humans should remain unchanged, (2) semantic similarity---the crafted example should bear the same meaning as the source, as judged by humans, and (3) language fluency---generated examples should look natural and grammatical. Previous works barely conform to all three requirements. For example, methods such as word misspelling~\cite{li2018textbugger,gao2018black}, single-word erasure~\cite{li2016understanding}, and phrase insertion and removal~\cite{liang2017deep} result in unnatural sentences. Moreover, there is almost no work that attacks the newly-risen BERT model on text classification. 
% While existing works on adversarial examples have showed success in the image and speech domain \cite{szegedy2013intriguing,carlini2018audio}, it is still challenging to deal with text data due to its discrete nature. 

% One reason is that words in the text are discrete tokens, and it is not possible to compute the gradient of the model loss function with respect to the input words, which hinders the use of many popular gradient-based methods. Another reason is that any small perturbations of text are indeed easily perceptible, even the replacement of a single word could drastically alter the semantics of the sentence.

% , which attracts human attention while reading, and can be easily detected by any spell checker. 

In this work, we present \textsc{TextFooler}, a simple but strong baseline for natural language attack in the black-box setting, a common case where no model architecture or parameters are accessible (Figure\ref{fig:intro-ex}. We design a more comprehensive paradigm to create both semantically and syntactically similar adversarial examples that meet the three desiderata mentioned above. 
% They attack CNN~\cite{liang2017deep, li2018textbugger, gao2018black} and LSTM~\cite{li2016understanding, li2018textbugger, gao2018black}.
% by relying on linguistic variation and vocabulary diversity, which does not make the sentence different to humans but can cause errors to many models. 
Specifically, we first identify the important words for the target model and then prioritize to replace them with the most semantically similar and grammatically correct words until the prediction is altered. We successfully applied this framework to attack three state-of-the-art models in five text classification tasks and two textual entailment tasks, respectively. On the adversarial examples, we can reduce the accuracy of almost all target models in all tasks to below $10\%$ with only less than $20\%$ of the original words perturbed. In addition, we validate that the generated examples are (1) correctly classified by human evaluators, (2) semantically similar to the original text, and (3) grammatically acceptable by human judges.

Our main contributions are summarized as follows:

\begin{itemize}
    \item We propose a simple but strong baseline, \textsc{TextFooler}, to quickly generate high-profile utility-preserving adversarial examples that force the target models to make wrong predictions under the black-box setting.
    \item We evaluate \textsc{TextFooler} on three state-of-the-art deep learning models over five popular text classification tasks and two textual entailment tasks, and it achieved the state-of-the-art attack success rate and perturbation rate.  
    \item We propose a comprehensive four-way automatic and three-way human evaluation of language adversarial attacks to evaluate the effectiveness, efficiency, and utility-preserving properties of our system.
    \item We open-source the code, pre-trained target models, and test samples for the convenience of future benchmarking.
    
\end{itemize}

\section{Method}

\subsection{Problem Formulation}

Given a corpus of $N$ sentences $\mathcal{X}= \{X_1, X_2,\dots, X_N\}$, and a corresponding set of $N$ labels $\mathcal{Y}=\{Y_1, Y_2,\dots, Y_N$\}, we have a pre-trained model $F: \mathcal{X} \to \mathcal{Y}$, which maps the input text space $\mathcal{X}$ to the label space $\mathcal{Y}$. 

For a sentence $X \in \mathcal{X}$, a valid adversarial example $X_{\mathrm{adv}}$ should conform to the following requirements: 
\begin{align}\label{eq:ad_requirement}
    F(X_{\mathrm{adv}}) \neq F(X), \text{and } 
    {\mathrm{Sim}}(X_{\mathrm{adv}}, X) \geq \epsilon
,
\end{align}
where $\mathrm{Sim}:\mathcal{X}\times\mathcal{X}\to (0,1)$ is a similarity function and $\epsilon$ is the minimum similarity between the original and adversarial examples.
In the natural language domain, $\mathrm{Sim}(\cdot)$ is often a semantic and syntactic similarity function. 

\subsection{Threat Model}

Under the black-box setting, the attacker is \textit{not} aware of the model architecture, parameters, or training data. It can only query the target model
with supplied inputs, getting as results the predictions and corresponding confidence scores. 

The proposed approach for adversarial text generation is shown in Algorithm \ref{alg:attack}, and consists of the two main steps:

% \paragraph{Step 1: Sentence importance ranking.}

% In the cases where the text example contains multiple sentences, usually the key opinion can be expressed by a few key sentences which play a more important role in the classification task. To improve the efficiency, we first rank the importance of each sentence to the final prediction results and then prioritize to manipulate them. Suppose we have a text example $x=(s_1,s_2,...,s_n)$, where $s_i$ is the $i$th sentence and $\mathcal{F}(x)=y$, we input each sentence into the target model and fetch the confidence score of label $y$ as the importance score of that sentence, i.e., $C_{s_i}=\mathcal{F}_y(s_i)$. To further reduce the computation burden, we remove those sentences that contribute negatively to the final prediction results, i.e., filter out sentences that have $\mathcal{F}(s_i)\neq y$.

\subsubsection{Step 1: Word Importance Ranking (line 1-6)}

Given a sentence of $n$ words $X=\{w_1, w_2, \dots, w_n\}$, we observe that only some keywords act as influential signals for the prediction model $F$, echoing with the discovery of \cite{niven2019probing} that BERT attends to the statistical cues of some words. Therefore, 
we create a selection mechanism to choose the words that most significantly influence the final prediction results. Using this selection process, we minimize the alterations, and thus maintain the semantic similarity as much as possible. 

Note that the selection of important words is trivial in a white-box scenario, as it can be easily solved by inspecting the gradients of the model $F$, while most other words are irrelevant. However, under the more common black-box set up in our paper, the model gradients are unavailable. Therefore, we create a selection mechanism as follows. We use the score $I_{w_i}$ to measure the influence of a word $w_i \in X$ towards the classification result $F(X)=Y$. We denote the sentence after deleting the word $w_i$ as $X_{\setminus w_i} = X\setminus \{w_i \}=\{w_1, \dots, w_{i-1}, w_{i+1},\dots w_n\}$, and use $F_Y(\cdot)$ to represent the prediction score for the $Y$ label. 

The importance score $I_{w_i}$ is therefore calculated as the prediction change before and after deleting the word $w_i$, which is formally defined as follows,
\begin{algorithm}[t]
\small
\caption{Adversarial Attack by \textsc{TextFooler}}
\label{alg:attack}
\begin{algorithmic}[1]

\REQUIRE{Sentence example $X=\{ w_1,w_2,...,w_n\}$, the corresponding ground truth label $Y$, target model $F$, sentence similarity function $\mathrm{Sim}(\cdot )$, sentence similarity threshold $\epsilon$}, word embeddings $\mathrm{Emb}$ over the vocabulary $\mathrm{Vocab}$.
\ENSURE{Adversarial example $X_{\mathrm{adv}}$}
\STATE Initialization: $X_{\mathrm{adv}}\leftarrow X$
\FOR{each word $w_i$ in $X$}
\STATE Compute the importance score $I_{w_i}$ via Eq.~\eqref{eq:words_importance}
\ENDFOR
\STATE 
\STATE Create a set $W$ of all words $w_i \in X$ sorted by the descending order of their importance score $I_{w_i}$.
\STATE Filter out the stop words in $W$.

\FOR{each word $w_j$ in $W$}
\STATE Initiate the set of candidates $\textsc{Candidates}$ by extracting the top $N$ synonyms using CosSim($\mathrm{Emb}_{w_j}$, $\mathrm{Emb}_{\mathrm{word}}$) for each word in $\mathrm{Vocab}$.
\STATE $\textsc{Candidates} \leftarrow \mathrm{POSFilter}(\textsc{Candidates})$
\STATE $\textsc{FinCandidates} \leftarrow \{ \text{ } \}$

\FOR{$c_{k}$ in $\textsc{Candidates}$}
\STATE $X'\leftarrow$ Replace $w_j$ with $c_{k}$ in $X_{\mathrm{adv}}$

\IF{$\mathrm{Sim}(X',X_{\mathrm{adv}}) >\epsilon$}
\STATE Add $c_k$ to the set \textsc{FinCandidates}
\STATE $Y_k \leftarrow F(X')$
\STATE $P_k \leftarrow F_{Y_k}(X')$
\ENDIF
\ENDFOR
\IF{there exists $c_k$ whose prediction result $Y_k \neq Y$}
\STATE In $\textsc{FinCandidates}$, only keep the candidates $c_k$ whose prediction result $Y_k \neq Y$
\STATE $c^*\leftarrow \argmax\limits_{c \in \textsc{FinCandidates}} \mathrm{Sim}(X, X'_{w_j \rightarrow c})$
\STATE $X_{\mathrm{adv}} \leftarrow$ Replace $w_j$ with $c^*$ in $X_{\mathrm{adv}}$
\RETURN $X_{\mathrm{adv}}$
\ELSIF{$P_{Y_k}(X_{\mathrm{adv}}) >\min\limits_{c_k \in \textsc{FinCandidates}}P_k $}
\STATE $c^*\leftarrow \argmin\limits_{c_k \in \textsc{FinCandidates}}P_k$
\STATE $X_{\mathrm{adv}} \leftarrow$ Replace $w_j$ with $c^*$ in $X_{\mathrm{adv}}$
\ENDIF
\ENDFOR
\RETURN \texttt{None}
\end{algorithmic}
\end{algorithm}
\begin{equation}
\small
\label{eq:words_importance}
    I_{w_i}=
    \begin{cases}
      F_Y(X)-F_Y(X_{\setminus w_i}),\qquad \text{if}\ F(X)=F(X_{\setminus w_i})=Y \\
      (F_Y(X)-F_Y(X_{\setminus w_i}))+(F_{\bar Y}(X_{\setminus w_i})-F_{\bar Y}(X)),\\\qquad\text{if}\ F(X)=Y, F(X_{\setminus w_i})=\bar Y,\text{ and } Y\neq \bar Y
      .
    \end{cases}
\end{equation}

After ranking the words by their  importance score, we further filter out stop words derived from NLTK\footnote{\url{https://www.nltk.org/}} and spaCy\footnote{\url{https://spacy.io/}} libraries such as ``the'', ``when'', and ``none''. This simple step of filtering is important to avoid grammar destruction.

\subsubsection{Step 2: Word Transformer (line 7-30)} 

Given the word $w_i \in X$ with a high importance score obtained in Step 1, we need to design a word replacement mechanism. A suitable replacement word needs to fulfill the following criteria: it should (1) have similar semantic meaning with the original one, (2) fit within the surrounding context, and (3) force the target model to make wrong predictions. In order to select replacement words that meet such criteria, we propose the following workflow.

\textbf{\textit{Synonym Extraction:}}  We gather a candidate set $\textsc{Candidates}$ for all possible replacements of the selected word $w_i$. $\textsc{Candidates}$ is initiated with $N$ closest synonyms according to the cosine similarity between $w_i$ and every other word in the vocabulary. 

To represent the words, we use word embeddings from \cite{mrkvsic2016counter}. These word vectors are specially curated for finding synonyms, as they achieve the state-of-the-art performance on SimLex-999, a dataset designed to measure how well different models judge the semantic similarity between words \cite{hill2015simlex}. 

Using this set of embedding vectors, we identify the top $N$ synonyms whose cosine similarity with $w$ is higher than $\delta$. Note that enlarging $N$ or lowering $\delta$ would both generate more diverse synonym candidates; however, the semantic similarity between the adversary and the original sentence would decrease. In our experiments, empirically setting $N$ to be 50 and $\delta$ to be $0.7$ strikes a balance between diversity and semantic similarity control.

\textbf{\textit{POS Checking:}} In the set $\textsc{Candidates}$ of the word $w_i$, we only keep the ones with the same part-of-speech (POS)\footnote{We used the off-the-shelf spaCy tagger, available at https://spacy.io/api/tagger} as $w_i$. This step is to assure that the grammar of the text is mostly maintained (line 10 in Algorithm \ref{alg:attack}).

\textbf{\textit{Semantic Similarity Checking:}} For each remaining word $c \in \textsc{Candidates}$, we substitute it for $w_i$ in the sentence $X$, and obtain the adversarial example $X_{\mathrm{adv}} = \{w_1, \dots, w_{i-1}, c ,w_{i+1}, \dots, w_{n}\}$.
We use the target model $F$ to compute the corresponding prediction scores $F(X_{\mathrm{adv}})$. We also calculate the sentence semantic similarity between the source $X$ and adversarial counterpart $X_{\mathrm{adv}}$. Specifically, we use Universal Sentence Encoder (USE) ~\cite{cer2018universal} to encode the two sentences into high dimensional vectors and use their cosine similarity score as an approximation of semantic similarity. The words resulting in similarity scores above a preset threshold $\epsilon$ are placed into the final candidate pool $\textsc{FinCandidates}$ (line 11-19 in Algorithm \ref{alg:attack}).

\textbf{\textit{Finalization of Adversarial Examples:}} In the final candidate pool $\textsc{FinCandidates}$, if there exists any candidate that can already alter the prediction of the target model, then we choose the word with the highest semantic similarity score among these winning candidates. But if not, then we select the word with the least confidence score of label $y$ as the best replacement word for $w_i$, and repeat Step 2 to transform the next selected word (line 20-30 in Algorithm \ref{alg:attack}).

Overall, the algorithm first uses Step 1 to rank the words by their importance scores, and then repeats Step 2 to find replacements for each word in the sentence $X$ until the prediction of the target model is altered.

\section{Experiments}
\label{sec:experiments}
\subsection{Tasks}
We study the effectiveness of our adversarial attack on two important NLP tasks, text classification and textual entailment.\footnote{Our datasets are at \url{http://bit.ly/nlp_adv_data}} The dataset statistics are summarized in Table \ref{tab:dataset_stats}. Following the practice by \citet{alzantot2018generating}, we evaluate our algorithm on a set of 1,000 examples randomly selected from the test set.

\begin{table}[ht]
\centering
\small
\begin{tabular}{c|l|ccc} 
    \Xhline{2\arrayrulewidth}
      \textbf{Task} & \textbf{Dataset}  & \textbf{Train}  & \textbf{Test} & \textbf{Avg Len}\\ \hline
      \multirow{ 5}{*}{\textbf{ Classification}} &AG's News &  120K  & 7.6K  & 43 \\
      &Fake News  & 18.8K & 2K & 885 \\
      &MR & 9K & 1K & 20  \\
      &IMDB & 25K & 25K &215 \\
      &Yelp& 560K & 38K & 152  \\
      \hline
      \multirow{ 2}{*}{\textbf{Entailment}} &SNLI& 570K &  3K & 8 \\
      &MultiNLI& 433K  &  10K & 11 \\ 
      \Xhline{2\arrayrulewidth}
    \end{tabular}
    \caption{Overview of the datasets.}
    \label{tab:dataset_stats}
\end{table}

\subsubsection{Text Classification}
To study the robustness of our model, we use text classification datasets with various properties, including news topic classification, fake news detection, and sentence- and document-level sentiment analysis, with average text length ranging from tens to hundreds of words. 
\begin{itemize}
    \item
    \textbf{AG's News (AG)}: Sentence-level classification with regard to four news topics: World, Sports, Business, and Science/Technology. 
    % The dataset contains $30,000$ news articles in the training set and $1,900$ in the test set~\cite{zhang2015character}. 
    Following the practice of \citet{zhang2015character}, we concatenate the title and description fields for each news article.
    
    \item
    \textbf{Fake News Detection (Fake)}: Document-level classification on whether a news article is fake or not. The dataset comes from the Kaggle Fake News Challenge.\footnote{\url{https://www.kaggle.com/c/fake-news/data}} We concatenate the title and news body of each article.% We split the dataset into a training set of $18,800$ samples and a test set of $2,000$ samples. 

    \item \textbf{MR}: Sentence-level sentiment classification on positive and negative movie reviews~\cite{pang2005seeing}. 
    % The dataset contains 5,331 positive and 5,331 negative reviews. 
    We use $90\%$ of the data as the training set and $10\%$ as the test set, following the practice in~\cite{li2018textbugger}.
    
    \item \textbf{IMDB}: Document-level sentiment classification on positive and negative movie reviews.\footnote{\url{https://datasets.imdbws.com/}} % The dataset has $25,000$ reviews in the training set and $25,000$ reviews in the test set. 
    
    \item \textbf{Yelp Polarity (Yelp)}: Document-level sentiment classification on positive and negative reviews~\cite{zhang2015character}. Reviews with a rating of 1 and 2 are labeled negative and 4 and 5 positive. 
    % The training set has $560,000$ samples and test set $38,000$ samples.

\end{itemize}

\subsubsection{Textual Entailment}
\begin{itemize}
    \item \textbf{SNLI}: A dataset of 570K sentence pairs derived from image captions. The task is to judge the relationship between two sentences: whether the second sentence can be derived from entailment, contradiction, or neutral relationship with the first sentence~\cite{snli:emnlp2015}.
    \item \textbf{MultiNLI}: A multi-genre entailment dataset with coverage of transcribed speech, popular fiction, and government reports~\cite{williams2017broad}. Compared to SNLI, it contains more linguistic complexity with various written and spoken English text.
\end{itemize}

\subsection{Attacking Target Models}

For each dataset, we train three state-of-the-art models on the training set, and achieved test set accuracy scores similar to the original implementation, as shown in Table~\ref{tab:tar_acc}. We then generate adversarial examples that are semantically similar to the test set to attack the trained models and make them generate different results.

\begin{table}[ht!]
\centering
\small
\begin{tabular}{rccccc} 
    \toprule

          & \textbf{WordCNN}  & \textbf{WordLSTM} & \textbf{BERT}\\ \hline
      \textbf{AG} &  92.5  & 93.1  & 94.6\\
      \textbf{Fake}  & 99.9 & 99.9 & 99.9\\
      \textbf{MR} & 79.9 & 82.2 & 85.8 \\
      \textbf{IMDB} & 89.7 & 91.2 &92.2\\
      \textbf{Yelp}& 95.2 & 96.6 & 96.1 \\
      \midrule
       & \textbf{InferSent}  & \textbf{ESIM} & \textbf{BERT}\\ \hline
     \textbf{SNLI} & 84.6 &  88.0 & 90.7\\
      \textbf{MultiNLI}& 71.1/71.5  &  76.9/76.5 & 83.9/84.1\\ 
      \toprule
    \end{tabular}
    \caption{Original accuracy of target models on standard test sets.}
    \label{tab:tar_acc}
\end{table}

On the sentence classification task, we target at three models: word-based convolutional neural network (WordCNN) \cite{kim2014convolutional}, word-based long-short term memory (WordLSTM) \cite{hochreiter1997long}, and the state-of-the-art Bidirectional Encoder Representations from
Transformers (BERT)~\cite{devlin2018bert}. 

% \FloatBarrier
\begin{table*}[!ht]
\small
\resizebox{\textwidth}{!}{\begin{tabular}{lccccc|ccccc|ccccc}
\toprule
& \multicolumn{5}{c|}{\textbf{WordCNN}} & \multicolumn{5}{c|}{\textbf{WordLSTM}} & \multicolumn{5}{c}{\textbf{BERT}} \\ 
& \textbf{MR}                   & \textbf{IMDB}                 & \textbf{Yelp}                 & \textbf{AG}                   & \textbf{Fake}                 & \textbf{MR}                   & \textbf{IMDB}                 & \textbf{Yelp}                 & \textbf{AG}                   & \textbf{Fake}                 & \textbf{MR}                   & \textbf{IMDB}                 & \textbf{Yelp}                 & \textbf{AG}                   & \textbf{Fake}                 \\ \hline
\begin{tabular}[c]{@{}l@{}}\textbf{Original Accuracy} \end{tabular}  &78.0&89.2&93.8&91.5&96.7&80.7&89.8 &96.0&91.3&94.0&86.0&90.9&97.0&94.2& 97.8                   \\ 
\begin{tabular}[c]{@{}l@{}}\textbf{After-Attack Accuracy} \end{tabular}   &2.8 & 0.0                     &1.1&1.5& 15.9&  3.1                    &        0.3              &2.1    &3.8 &16.4 & 11.5  &  13.6  &6.6&12.5&19.3      \\ 
\begin{tabular}[c]{@{}l@{}} \textbf{\% Perturbed Words} \end{tabular}&14.3 & 3.5&8.3&15.2&11.0 &  14.9                    &5.1 &10.6& 18.6 &10.1 &16.7&6.1&13.9&22.0& 11.7                     \\ 
\begin{tabular}[c]{@{}l@{}}\textbf{Semantic Similarity} \end{tabular}      &0.68&0.89&0.82&0.76&0.82&0.67&0.87&0.79&0.63 &0.80&0.65 &0.86&0.74&0.57 &  0.76                    \\
\begin{tabular}[c]{@{}l@{}} \textbf{Query Number} \end{tabular}   &123  &  524&487&228&3367 & 126                     &666&629  & 273 &3343&166  &1134&827  &357& 4403     \\ 
 \hline
\begin{tabular}[c]{@{}l@{}} \textbf{Average Text Length} \end{tabular}       &  \multicolumn{1}{c}{20} & \multicolumn{1}{c}{215} & \multicolumn{1}{c}{152} & \multicolumn{1}{c}{43} & \multicolumn{1}{c|}{885} & \multicolumn{1}{c}{20} & \multicolumn{1}{c}{215} & \multicolumn{1}{c}{152} & \multicolumn{1}{c}{43} & \multicolumn{1}{c|}{885} & \multicolumn{1}{c}{20} & \multicolumn{1}{c}{215} & \multicolumn{1}{c}{152} & \multicolumn{1}{c}{43} & \multicolumn{1}{c}{885} \\ 
\bottomrule
\end{tabular}}
\caption{Automatic evaluation results of the attack system on text classification datasets, including the original model prediction accuracy before being attacked (``Original Accuracy''), the model accuracy after the adversarial attack (``After-Attack Accuracy''), the percentage of perturbed words with respect to the original sentence length (``\% Perturbed Words''), and the semantic similarity between original and adversarial samples (``Semantic Similarity'').}
\label{table:class-auto-eval}
\end{table*}
% \FloatBarrier

For the WordCNN model, we used three window sizes of 3, 4, and 5, and 100 filters for each window size with a dropout of $0.3$.
For the WordLSTM, we used a 1-layer bidirectional LSTM with $150$ hidden units and a dropout of $0.3$. For both models, we used the 200-dimensional Glove word embeddings pre-trained on 6B tokens from Wikipedia and Gigawords \cite{pennington2014glove}. 
We used the 12-layer BERT model with $768$ hidden units and $12$ heads, with $110$M parameters, which is called the base-uncased version.\footnote{\url{https://github.com/huggingface/pytorch-pretrained-BERT}}

We also implemented three target models on the textual entailment task: standard InferSent\footnote{\url{https://github.com/facebookresearch/InferSent}}~\cite{conneau2017supervised}, ESIM\footnote{\url{https://github.com/coetaur0/ESIM}}~\cite{chen2016enhanced}, and fine-tuned BERT.

\subsection{Setup of Automatic Evaluation}

We first report the accuracy of the target models on the original test samples before the attack as the original accuracy. Then we measure the accuracy of the target models against the adversarial samples crafted from the test samples, denoted as \textit{after-attack accuracy}. By comparing these two accuracy scores, we can evaluate how successful the attack is, --- the more significant the gap between the original and after-attack accuracy signals, the more successful our attack is. Apart from these accuracies, we also report the perturbed word percentage as the ratio of the number of perturbed words to the text length. Furthermore, we apply USE\footnote{https://tfhub.dev/google/
universal-sentence-encoder} to measure the semantic similarity between the original and adversarial texts. These two metrics, the perturbed words percentage and the semantic similarity score, which together evaluate how semantically similar the original and adversarial texts are. We finally report the number of queries the attack system made to the target model and fetch the output probability scores. This metric can reveal the efficiency of the attack model.

\begin{table*}[ht]
\small
\centering

\begin{tabular}{lcc|cc|cc}
\toprule
 & \multicolumn{2}{c|}{\textbf{InferSent}}               & \multicolumn{2}{c|}{\textbf{ESIM}}                    & \multicolumn{2}{c}{\textbf{BERT}}                    \\ 
 & \textbf{SNLI}                 & \textbf{MultiNLI (m/mm)}      & \textbf{SNLI}                 & \textbf{MultiNLI (m/mm)}      & \textbf{SNLI}                 & \textbf{MultiNLI (m/mm)}      \\ \hline
\begin{tabular}[c]{@{}l@{}}\textbf{Original Accuracy} \end{tabular} & 84.3  & 70.9/69.6  & 86.5 & 77.6/75.8 & 89.4& 85.1/82.1                      \\ 
\begin{tabular}[c]{@{}l@{}}\textbf{After-Attack Accuracy} \end{tabular}   &    3.5                  &  6.7/6.9      &5.1 & 7.7/7.3  &4.0 &9.6/8.3\\ 
\begin{tabular}[c]{@{}l@{}}\textbf{\% Perturbed Words}\end{tabular}    &    18.0                  &  13.8/14.6       &18.1 &14.5/14.6&18.5  &15.2/14.6  \\ 
\begin{tabular}[c]{@{}l@{}}\textbf{Semantic Similarity}\end{tabular}    &0.50&0.61/0.59&0.47&0.59/0.59&0.45&0.57/0.58                     \\
\begin{tabular}[c]{@{}l@{}}\textbf{Query Number}\end{tabular}      &   57& 70/83&58 &72/87& 60& 78/86 \\ 
 \hline
\begin{tabular}[c]{@{}l@{}}\textbf{Average Text Length}\end{tabular}     & \multicolumn{1}{c}{8} & \multicolumn{1}{c|}{11/12} & \multicolumn{1}{c}{8} & \multicolumn{1}{c|}{11/12} & \multicolumn{1}{c}{8} & \multicolumn{1}{c}{11/12} \\ 
\bottomrule
\end{tabular}
\caption{Automatic evaluation results of the attack system on textual entailment datasets. ``m'' means matched, and ``mm'' means mismatched, which are the two variants of the MultiNLI development set.}
\label{table:nli-auto-eval}
\end{table*}

\subsection{Setup of Human Evaluation}

We conduct human evaluation on three criteria: semantic similarity, grammaticality, and classification accuracy. 
% We aim to generate sentences that are clearly similar to the original test set in the eyes of humans but confuses the neural models. 
We randomly select 100 test sentences of each task to generate adversarial examples, one targeting WordLSTM on the MR dataset and another targeting BERT on SNLI. We first shuffled a mix of original and adversarial texts and asked human judges to rate the grammaticality of them on a Likert scale of $1-5$, similar to the practice of~\cite{gagnon2018salsa}. Next, we evaluate the classification consistency by asking humans to classify each example in the shuffled mix of the original and adversarial sentences and then calculate the consistency rate of both classification results. Lastly, we evaluated the semantic similarity of the original and adversarial sentences by asking humans to judge whether the generated adversarial sentence is similar, ambiguous, or dissimilar to the source sentence. Each task is completed by two independent human judges who are native English speakers. The volunteers have university-level education backgrounds and passed a test batch before they started annotation.

% \subsection{Experimental Details}

% In experiments, we found that when the text length is less than 10 words, any change of a single word would dramatically change the semantic similarity score computed by sentence encoding. 

\section{Results}

\subsection{Automatic Evaluation}

The main results of black-box attacks  in terms of automatic evaluation on five text classification and two textual entailment tasks are summarized in Table \ref{table:class-auto-eval} and \ref{table:nli-auto-eval}, respectively. Overall, as can be seen from our results, \textsc{TextFooler} achieves a high success rate when attacking with a limited number of modifications on both tasks. No matter how long the text sequence is, and no matter how accurate the target model is, \textsc{TextFooler} can always reduce the accuracy from the state-of-the-art values to below 15\% (except on the Fake dataset) with less than 20\% word perturbation ratio (except the AG dataset under the BERT target model). For instance, it only perturbs 5.1\% of the words on average when reducing the accuracy from 89.8\% to only 0.3\% on the IMDB dataset against the WordLSTM model. Notably, our attack system makes the WordCNN model on the IMDB dataset totally wrong (reaching the accuracy of 0\%) with only 3.5\% word perturbation rate. In the IMDB dataset, which has an average length of 215 words, the system only perturbed 10 words or fewer per sample to conduct successful attacks. This means that our attack system can successfully mislead the classifiers into assigning wrong predictions via subtle manipulation.

Even for BERT, which has achieved seemingly ``robust'' performance compared with the non-pretrained models such as WordLSTM and WordCNN, our attack model can still reduce its prediction accuracy by about 5--7 times on the classification task (e.g., from 95.6\% to 6.8\% for Yelp dataset) and about 9-22 times on the NLI task (e.g., from 89.4\% to 4.0\% for SNLI dataset), which is unprecedented. Our curated adversarial examples can contribute to the study of the interpretability of the BERT model \cite{feng2018pathologies}.

Another two observations can be drawn from Table \ref{table:class-auto-eval} and \ref{table:nli-auto-eval}. (1) Models with higher original accuracy are, in general, more difficult to be attacked. For instance, the after-attack accuracy and perturbed word ratio are both higher for the BERT model compared with WordCNN on all datasets. (2) The after-attack accuracy of the Fake dataset is much higher than all other classification datasets for all three target models. We found in experiments that it is easy for the attack system to convert a real news article to a fake one, whereas the reverse process is much harder, which is in line with intuition.

% For both tasks, when the target models get more accurate, the attack accuracy also increases and the perturbation rate gets higher. For instance, for the MultiNLI dataset, the original accuracy of target models keep increasing from InferSent to ESIM and then to BERT, so do their attack accuracy and perturbed words ratio by looking at Table \ref{table:nli-auto-eval}. This indicates that the attack model needs to change more parts of the text to fool a wiser target model, which conforms to intuition. To be noted, for the five text classification datasets, although the original accuracy of the BERT model have not been elevated by much compared with the WordCNN and WordLSTM models, the attack accuracy have been boosted significantly. This indicates that the generalization and robustness of BERT model is much better than normal models that do no involve large-scale pre-training. 

Comparing the semantic similarity scores and the perturbed word ratios in both Table \ref{table:class-auto-eval} and \ref{table:nli-auto-eval}, we find that the two results have a high positive correlation. Empirically, when the text length is longer than 10 words, the semantic similarity measurement becomes more stable. Since the average text lengths of text classification datasets are all above 20 words and those of textual entailment datasets are around or below 10 words, we need to treat the semantic similarity scores of these two tasks individually. Therefore, we performed a linear regression analysis between the word perturbation ratio and semantic similarity for each task and obtained r-squared values of 0.94 and 0.97 for text classification and textual entailment tasks, respectively. Such high values of r-squared reveal that our proposed semantic similarity has a high (negative) correlation with the perturbed words ratio, which can both be useful automatic measurements to evaluate the degree of alterations of the original text.  

We include the average text length of each dataset in the last row of Table \ref{table:class-auto-eval} and \ref{table:nli-auto-eval} so that it can be conveniently compared against the query number. The query number is almost linear to the text length, with a ratio in $(2,8)$. Longer text correlates with a smaller ratio, which validates the efficiency of \textsc{TextFooler}.

\paragraph{Benchmark Comparison} We compared \textsc{TextFooler} with the previous state-of-the-art adversarial attack systems against the same target model and dataset. Our baselines include \cite{li2018textbugger} that generates misspelled words by character- and word-level perturbation, \cite{alzantot2018generating} that iterates through every word in the sentence and find its perturbation, and \cite{kuleshov2018adversarial} that uses word replacement by greedy heuristics. From the results in Table \ref{table:benchmark}, we can see that our system beats the previous state-of-the-art models by both the attack success rate (calculated by dividing the number of wrong predictions by the total number of adversarial examples) and perturbed word ratio.

\begin{table}[]
\small
\centering
\resizebox{1.\columnwidth}{!}{\begin{tabular}{llcc}
\toprule
\textbf{Dataset}               & \textbf{Model} & \textbf{Success Rate} & \textbf{\% Perturbed Words} \\ \hline
\multirow{3}{*}{\textbf{IMDB}} & \cite{li2018textbugger}      & 86.7         & 6.9            \\ 
                      & \cite{alzantot2018generating} & 97.0& 14.7\\ 
                      & Ours  & \textbf{99.7}         & \textbf{5.1}    \\ \hline
\multirow{2}{*}{\textbf{SNLI}} & \cite{alzantot2018generating}      & 70.0         & 23.0  \\  
                      & Ours  & \textbf{95.8}         & \textbf{18.0}           \\ \hline
\multirow{2}{*}{\textbf{Yelp}} & \cite{kuleshov2018adversarial}      & 74.8         & -      \\  
                      & Ours  & \textbf{97.8}         & \textbf{10.6}           \\ \bottomrule
\end{tabular}
}
\caption{Comparison of our attack system against other published systems. The target model for IMDB and Yelp is LSTM and SNLI is InferSent.}
\label{table:benchmark}

\end{table}

\subsection{Human Evaluation}

\begin{table*}[!ht]
\centering
\small
\begin{tabular}{p{3.4cm}p{13.5cm}}
    \toprule

         \multicolumn{2}{c}{\textbf{Movie Review (Positive (POS) $\leftrightarrow$ Negative (NEG))}}  \\ \hline
       \textbf{Original (Label: NEG) }& The characters, cast in impossibly \textbf{\textit{contrived situations}}, are \textbf{\textit{totally}} estranged from reality. \\ 
      \textbf{Attack (Label: POS)} & The characters, cast in impossibly \textbf{\textit{engineered circumstances}}, are \textbf{\textit{fully}} estranged from reality. \\ \hline
      \textbf{Original (Label: POS)} & It cuts to the \textbf{\textit{knot}} of what it actually means to face your \textbf{\textit{scares}}, and to ride the \textbf{\textit{overwhelming}} \textbf{metaphorical wave} that life wherever it takes you. \\ 
      \textbf{Attack (Label: NEG)} & It cuts to the \textbf{\textit{core}} of what it actually means to face your \textbf{\textit{fears}}, and to ride the \textbf{\textit{big}}\textbf{ metaphorical wave} that life wherever it takes you. \\ 
      \midrule
      \multicolumn{2}{c}{\textbf{SNLI (Entailment (ENT), Neutral (NEU), Contradiction (CON))}} \\ \hline

      \textbf{Premise} & Two small boys in blue soccer uniforms use a wooden set of steps to wash their hands. \\ 
      \textbf{Original (Label: CON)} & The boys are in band \textbf{\textit{uniforms}}. \\ 
      \textbf{Adversary (Label: ENT)} & The boys are in band \textbf{\textit{garment}}. \\ \hline
      
      \textbf{Premise} & A child with wet hair is holding a butterfly decorated beach ball. \\ 
      \textbf{Original (Label: NEU)} & The \textbf{\textit{child}} is at the \textbf{\textit{beach}}. \\ 
      \textbf{Adversary (Label: ENT)} & The \textbf{\textit{youngster}} is at the \textbf{\textit{shore}}. \\
      
      \bottomrule
    \end{tabular}
    \caption{Examples of original and adversarial sentences from MR (WordLSTM) and SNLI (BERT) datasets.}
    \label{tab:eg}
\end{table*}
We sampled 100 adversarial examples on the MR dataset with the WordLSTM and 100 examples on SNLI with BERT. We verified the quality of our examples via three experiments. First, we ask human judges to give a grammaticality score of a shuffled mix of original and adversarial text.

%  Grammaticality is an essential criterion for adversarial examples because it does not make sense to generate gibberish English to confuse the model. 

\begin{table}[!ht]
\centering
\small
\begin{tabular}{lccc}
    \Xhline{2\arrayrulewidth}

         & \textbf{MR}  & \textbf{SNLI} \\
       \textbf{Source Text}  & \textbf{(WordLSTM)}  &  \textbf{(BERT)}\\ \hline
      \textbf{Original} & 4.22 & 4.50 \\
      
      \textbf{Adversarial} & 4.01  & 4.27  \\ 
      
    \Xhline{2\arrayrulewidth}
    \end{tabular}
    \caption{Grammaticality of original and adversarial examples for MR (WordLSTM) and SNLI (BERT) on $1-5$ scale.}
    \label{tab:hum_eval_gra}
\end{table}
As shown in Table~\ref{tab:hum_eval_gra}, the grammaticality of the adversarial text are close to the original text on both datasets. By sensibly substituting synonyms, \textsc{TextFooler} generates smooth outputs such as ``the big metaphorical wave'' in Table~\ref{tab:eg}. 
 
We then asked the human raters to assign classification labels to a shuffled set of original and adversarial samples. 
%  We showed a set of sentences with positive/negative labels for the sentiment classification on MR and sentence pairs with entailment/neutral/contradiction relationships for SNLI samples. 
%  Due to the nature of tasks, it is easy for humans to agree on the ground truth labels on sentiment analysis but harder on natural inference. Nonetheless, 
The overall agreement between the labels of the original sentence and the adversarial sentence is relatively high, with $92\%$ on MR and $85\%$ on SNLI. 
%  (in Table~\ref{tab:hum_eval_classi}). 
 Though our adversarial examples are not perfect in every case, this shows that majorities of adversarial sentences have the same attribute as the original sentences from humans' perspective. Table~\ref{tab:eg} shows typical examples of sentences with almost the same meanings that result in contradictory classifications by the target model.

% \begin{table}[!ht]
% \centering
% \small
% \begin{tabular}{lccc}
%     \toprule

%          & MR  & SNLI \\
%       Input  & (WordLSTM)  &  (BERT)\\ \hline
%       Original Accuracy (\%) &88  &    68 \\
      
%       Attack Accuracy (\%) & 82  & 51  \\ 
%       Agreement (\%)&  83  & 72 \\ 
      
%       \toprule
%     \end{tabular}
%     \caption{Human classification accuracy on adversarial examples for MR (WordLSTM) and SNLI (BERT) models.}
%     \label{tab:hum_eval_classi}
% \end{table}

Lastly, we asked the judges to determine whether each adversarial sample retains the meaning of the original sentence. They need to decide whether the synthesized adversarial example is similar, ambiguous, or dissimilar to the provided original sentence. We regard similar as $1$, ambiguous as $0.5$, and dissimilar as $0$, and obtained sentence similarity scores of $0.91$ on MR and $0.86$ on SNLI, which shows the perceived difference between original and adversarial text is small.
% The higher semantic similarity of adversarial and original examples from MR correlates with the lower percentage of perturbed words in MR.

\section{Discussion}

\subsection{Ablation Study}

\subsubsection{Word Importance Ranking}

To validate the effectiveness of Step 1 in Algorithm \ref{alg:attack}, i.e., the word importance ranking, we remove this step and instead randomly select the words in the text to perturb. We keep the perturbed word ratio and Step 2 the same. We use BERT as the target model and test on three datasets: MR, AG, and SNLI. The results are summarized in Table \ref{table:random-perturb}. After removing Step 1 and instead randomly selecting the words to perturb, the after-attack accuracy increases by more than $45\%$ on all three datasets, which reveals that the attack becomes ineffective without the word importance ranking step. The word importance ranking process is crucial to the algorithm in that it can accurately and efficiently locate the words which cast the most significant effect on the predictions of the target model. This strategy can also reduce the number of perturbed words so as to maintain the semantic similarity as much as possible.

\begin{table}[ht]
\centering
\small
\begin{tabular}{lccc}
\Xhline{2\arrayrulewidth}
                                                                        & \textbf{MR} & \textbf{AG} & \textbf{SNLI} \\ \hline
\begin{tabular}[c]{@{}l@{}}\textbf{\% Perturbed Words} \end{tabular}           &  16.7  &  22.0  & 18.5     \\ 
\begin{tabular}[c]{@{}l@{}}\textbf{Original Accuracy} \end{tabular}             & 86.0   & 94.2   & 89.4     \\
\begin{tabular}[c]{@{}l@{}}\textbf{After-Attack Accuracy} \end{tabular} & 11.5   &  12.5  & 4.0     \\ \hline
\begin{tabular}[c]{@{}l@{}}\textbf{After-Attack Accuracy (Random)}\end{tabular} &  \textbf{68.3}  & \textbf{80.8}  &   \textbf{59.2}   \\ \Xhline{2\arrayrulewidth}
\end{tabular}
\caption{Comparison of the after-attack accuracies before and after removing the word importance ranking of Algorithm \ref{alg:attack}. For control, Step 2 and the perturbed words ratio are kept the same. BERT model is used as the target model.}
\label{table:random-perturb}
\end{table}

\subsubsection{Semantic Similarity Constraint}

In Step 2 of Algorithm \ref{alg:attack}, for every possible word replacement, we check the semantic similarity between the newly generated sample and the original text, and adopt this replacement only when the similarity is above a preset threshold $\epsilon$. We found that this strategy can effectively filter out irrelevant synonyms to the selected word. As we can see from the examples in Table \ref{table:seman-sim}, the synonyms extracted by word embeddings are noisy, so directly injecting them into the text as adversarial samples would probably shift the semantic meaning significantly. By applying the sentence-level semantic similarity constraint, we can obtain more related synonyms as suitable replacements. Table \ref{table:seman-sim-ablation} compares the automatic evaluation metrics before and after removing the semantic similarity constraint for the BERT-Base target model. From this table, we see that without the constraint, the attacking becomes easier (after-attack accuracy, perturbed  words  percentage, and number of queries all decrease). However, the semantic similarity between the original text and the corresponding adversaries also decrease, indicating the degradation of meaning preservation.\footnote{To be noted, the decrease of semantic similarity score is much smaller for the NLI datasets. This is because the lengths of NLI sentences are generally less than 20 tokens, and the USE encoder is not as sensitive to the change of semantic meaning in this situation.}
% In future work, we will also explore the use of ELMo \cite{Peters:2018} or BERT \cite{devlin2018bert} to calculate the sentence similarity scores.

\begin{table}[ht]
\centering
\small
\begin{tabular}{p{1.4cm}p{6.3cm}}

\Xhline{2\arrayrulewidth}
\textbf{Original}                                                            & like a south of the border melrose \textbf{\textit{place}}          \\ 
\textbf{Adversarial}                                                            & like a south of the border melrose \textbf{\textit{spot}}           \\ 
\begin{tabular}[c]{@{}l@{}} \textbf{ \textit{ } - Sim.} \end{tabular} & like a south of the border melrose \textbf{\textit{mise}}           \\ \hline
\textbf{Original}                                                            & their computer animated faces are very \textbf{\textit{expressive}} \\ 
\textbf{Adversarial} & their computer animated face are very \textbf{\textit{affective}}   \\ 
\begin{tabular}[c]{@{}l@{}} \textbf{ \textit{ } - Sim.} \end{tabular} & their computer animated faces are very \textbf{\textit{diction}}    \\ \Xhline{2\arrayrulewidth}
\end{tabular}
\caption{Qualitative comparison of adversarial attacks with and without the semantic similarity constraint (``-Sim.''). We highlight the 
original word, TextFooler's replacement, and the replacement without semantic constraint.}
\label{table:seman-sim}
\end{table}

\begin{table}[htpb]
\centering
\small
\resizebox{1.\columnwidth}{!}{\begin{tabular}{lrrrr}
\Xhline{2\arrayrulewidth}
                      & \textbf{MR}        & \textbf{IMDB}      & \textbf{SNLI}      & \textbf{MNLI(m)}    \\ \hline
\textbf{After-Attack Accu.} & 11.5/6.2  & 13.6/11.2 & 4.0/3.6   & 9.6/7.9   \\ 
\textbf{\% Perturbed Words}    & 16.7/14.8 & 6.1/4.0   & 18.5/18.3 & 15.2/14.5 \\ 
\textbf{Query Number}          & 166/131   & 1134/884  & 60/57     & 78/70     \\ 
\textbf{Semantic Similarity}   & 0.65/0.58 & 0.86/0.82 & 0.45/0.44 & 0.57/0.56 \\ 
\Xhline{2\arrayrulewidth}
\end{tabular}}
\caption{Comparison of automatic evaluation metrics with and without the semantic similarity constraint. (Numbers in the left and right of the symbol ``/'' represent results with and without the constraint, respectively). The target model is BERT-Base.}
\label{table:seman-sim-ablation}
\end{table}

\subsection{Transferability}

We examined the transferability of adversarial text, that is, whether adversarial samples curated based on one model can also fool another. For this, we collected the adversarial examples from IMDB and SNLI test sets that are wrongly predicted by one target model and then measured the prediction accuracy of them against the other two target models. As we can see from the results in Table \ref{tab:transfer}, there is a moderate degree of transferability between models, and the transferability is higher in the textual entailment task than in the text classification task. Moreover, the adversarial samples generated based on the model with higher prediction accuracy, i.e., the BERT model here, show higher transferability.

\begin{table}[!ht]
\small
\centering
\resizebox{\columnwidth}{!}{\begin{tabular}{llccc}
\Xhline{2\arrayrulewidth}
 & & \textbf{WordCNN} & \textbf{WordLSTM} & \textbf{BERT} \\ \hline
\multirow{3}{*}{\textbf{IMDB}} & \textbf{WordCNN}  & ---     & 84.9     & 90.2 \\ 
& \textbf{WordLSTM} & 74.9    & ---      & 87.9 \\ 
& \textbf{BERT}     & 84.1    & 85.1     & ---  \\ 
\Xhline{2\arrayrulewidth}
        &  & \textbf{InferSent} & \textbf{ESIM} & \textbf{BERT} \\ \hline
\multirow{3}{*}{\textbf{SNLI}} & \textbf{InferSent} & ---       & 62.7 & 67.7 \\ 
& \textbf{ESIM}      & 49.4      & ---  & 59.3 \\ 
& \textbf{BERT}      & 58.2      & 54.6 & ---  \\ \Xhline{2\arrayrulewidth}
\end{tabular}}
\caption{Transferability of adversarial examples on IMDB and SNLI dataset. Row $i$ and column $j$ is the accuracy of adversaries generated for model $i$ evaluated on model $j$.}
\label{tab:transfer}
\end{table}

% \begin{table}[!ht]
% \small
% \centering
% \begin{tabular}{lccc}
% \hline
%           & InferSent & ESIM & BERT \\ \hline
% InferSent & 0.0       & 62.7 & 67.7 \\ 
% ESIM      & 49.4      & 0.0  & 59.3 \\ 
% BERT      & 58.2      & 54.6 & 0.0  \\ \hline
% \end{tabular}
% \caption{Transferability of adversarial examples on the SNLI dataset. Row i and column j show the accuracy of adversarial samples generated for model i evaluated on model j.}
% \label{tab:transfer-snli}
% \end{table}

\subsection{Adversarial Training}

Our work casts insights on how to better improve the original models through these adversarial examples. We conducted a preliminary experiment on adversarial training, by feeding the models both the original data and the adversarial examples (adversarial examples share the same labels as the original counterparts), to see whether the original models can gain more robustness. We collected the adversarial examples curated from the MR and SNLI training sets that fooled BERT and added them to the original training set. We then used the expanded data to train BERT from scratch and attacked this adversarially-trained model. As is seen in the attack results in Table \ref{tab:adv-training}, both the after-attack accuracy and perturbed words ratio after adversarial re-training get higher, indicating the greater difficulty to attack. This reveals one of the potencies of our attack system,--- we can enhance the robustness of a model to future attacks by training it with the generated adversarial examples.

\begin{table}[!ht]
\small
\centering
% \resizebox{\columnwidth}{!}{

\begin{tabular}{lcccc}
\Xhline{2\arrayrulewidth}
                  & \multicolumn{2}{c}{\textbf{MR}}                  & \multicolumn{2}{c}{\textbf{SNLI}}                \\ 
                  & \textbf{Af. Acc.} & \textbf{Pert.} & \textbf{Af. Acc.} & \textbf{Pert.} \\ \hline
\textbf{Original}  & 11.5             & 16.7                 & 4.0              & 18.5                 \\
\textbf{+ Adv. Training}   & \textbf{18.7}             & \textbf{21.0}                 & \textbf{8.3}              & \textbf{20.1}                 \\ \Xhline{2\arrayrulewidth}
\end{tabular}
% }
\caption{Comparison of the after-attack accuracy (``Af. Acc.'') and percentage of perturbed words (``Pert.'') of original training (``Original'') and adversarial training (``+ Adv. Train'') of BERT model on MR and SNLI dataset.}
\label{tab:adv-training}
\end{table}

\subsection{Error Analysis}
Our adversarial samples are susceptible to three types of errors: word sense ambiguity, grammatical error, and task-sensitive content shift. Although large thesauri are available, a word usually has many meanings, with a set of synonyms for each word sense. 
One example can be the transfer from an original sentence ``One man \textit{shows} the ransom money to the other'' to the synthesized ``One man \textit{testify} the ransom money to the other'', where ``testify'' in this case is not the appropriate synonym of ``show''.
Grammatical errors are also frequent in text generation. For example, the sentence ``A man with headphones is \textit{biking}'' and ``A man with headphones is \textit{motorcycle}'' differ by the word ``biking'', which can be both a noun and a verb, as well as a reasonably similar word to ``motorcycle''. 
% Some even more subtle grammatical error can be seen on adverbs, such as the pair ``A boy is sitting still'', vesus ``A boy is sitting anymore''. 

As future work, some carefully designed heuristics can be applied to filter out grammatical errors.
The content shift can be seen in a task-specific situation. In the sentiment classification task, a change of words might not affect the overall sentiment, whereas in the task of textual entailment, the substitution of words might result in a fundamental difference. For example, if the premise is ``a \textit{kid} with red hat is running'', and the original hypothesis is ``a \textit{kid} is running (\textit{entailment})'', then if the adversarial example becomes ``a \textit{girl} is running'', the sensible result turns into \textit{neutral} instead.

\section{Related Work}
Adversarial attack has been extensively studied in computer vision \cite{goodfellow2014explaining,kurakin2016adversarial,moosavi2017universal}.
Most works make gradient-based perturbation on continuous input spaces~\cite{szegedy2013intriguing,goodfellow2014explaining}.
Adversarial attack on discrete data such as text is more challenging. Inspired by the approaches in computer vision, early work in language adversarial attack focus on variations of gradient-based methods. For example, \citet{zhao2017generating} transform input data into a latent representation by generative adversarial networks (GANs), and then retrieved adversaries close to the original instance in the latent space. 
Other works observed the intractability of GAN-based models on the text and the shift in semantics in the latent representations, so heuristic methods such as scrambling, misspelling, or removing words were proposed~\cite{ebrahimi2017hotflip,alzantot2018generating,li2016understanding,li2018textbugger}. Instead of manually designing these rules, \citet{ribeiro2018semantically} automatically craft the semantically equivalent adversarial rules from the machine-generated paraphrases based on the back-translation technique from the machine translation field. By human evaluation, they demonstrate that the quality of the adversaries created by their rules is better than that generated by humans. More interestingly, \citet{feng2018pathologies} used the way to produce adversarial examples to help improve the interpretability of neural models by encouraging high entropy outputs on the adversaries. 
% For example, \cite{ebrahimi2017hotflip} adopted a heuristic to substitute single words adversarially. 
% However, such heuristic approaches rely on white-box access to the model.
\section{Conclusion}
Overall, we study adversarial attacks against state-of-the-art text classification and textual entailment models under the black-box setting. Extensive experiments demonstrate the effectiveness of our proposed system, \textsc{TextFooler}, at generating targeted adversarial texts. Human studies validated that the generated adversarial texts are legible, grammatical, and similar in meaning to the original texts.

\section{Acknowledgments}
We thank the MIT Clinical Decision-Making Group and Professor Zheng Zhang for insightful discussions. We especially appreciate Heather Berlin, Yilun Du, Laura Koemmpel, and many other helpers for their high-quality human evaluation. Zhijing Jin also thank Zhutian Yang and Yixuan Zhang for strong personal support.

\bibliographystyle{aaai}

\bibliography{references}

\end{document}